\documentclass[11pt,a4paper]{article}
\usepackage[hyperref]{emnlp20}
\usepackage{times}
\usepackage{latexsym}

\aclfinalcopy
\usepackage{microtype}
\usepackage{booktabs}       
\usepackage{amsfonts}
\usepackage{amsmath}
\usepackage{algorithm}
\usepackage{algorithmic}
\usepackage{url}
\usepackage{multirow}
\usepackage{graphics}
\usepackage{graphicx}
\usepackage{float}

\title{Pseudo-Bidirectional Decoding for Local Sequence Transduction}

\author{
Wangchunshu Zhou$^{1}$\thanks{\ \ This work was done during the first author's internship at Microsoft Research Asia.} ~~~ Tao Ge$^{2}$ ~~~ Ke Xu$^{1}$ \\
$^1$Beihang University, Beijing, China\\
$^2$Microsoft Research Asia, Beijing, China\\
{\tt zhouwangchunshu@buaa.edu.cn, kexu@nlsde.buaa.edu.cn}\\
{\tt tage@microsoft.com}\\}

\begin{document}

\maketitle

\begin{abstract}
Local sequence transduction (LST) tasks are sequence transduction tasks where there exists massive overlapping between the source and target sequences, such as grammatical error correction and spell or OCR correction. Motivated by this characteristic of LST tasks, we propose Pseudo-Bidirectional Decoding (PBD), a simple but versatile approach for LST tasks. PBD copies the representation of source tokens to the decoder as pseudo future context that enables the decoder self-attention to attends to its bi-directional context. In addition, the bidirectional decoding scheme and the characteristic of LST tasks motivate us to share the encoder and the decoder of LST models. Our approach provides right-side context information for the decoder, reduces the number of parameters by half, and provides good regularization effects. Experimental results on several benchmark datasets show that our approach consistently improves the performance of standard seq2seq models on LST tasks.
\end{abstract}
\section{Introduction}
\begin{figure}
    \centering
    \includegraphics[width=\linewidth]{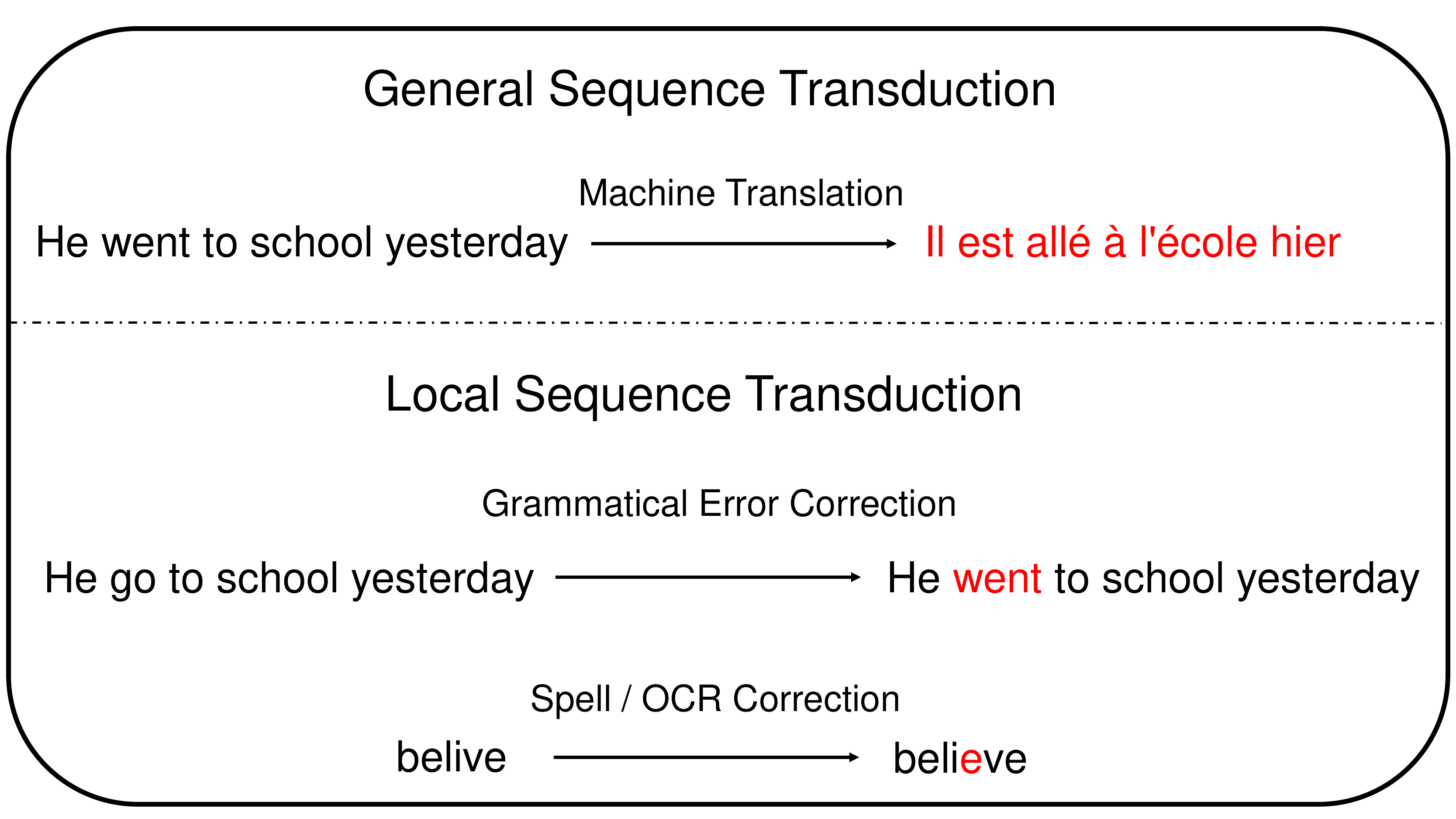}
    \caption{Illustration of the characteristic of local transduction tasks versus general sequence transduction tasks. Words and letters in red are those different from that in the input sequences.}
    \label{example}
\end{figure}

As illustrated in Figure \ref{example}, in local sequence transduction (LST) tasks, a model is trained to map an input sequence $x_{1}, . . . , x_{n}$ to an output sequence $y_{1}, . . . , y_{m}$, where the input and output sequences are of similar length and differ only in a few positions. Many important NLP tasks can be formulated as LST tasks, including automatic grammatical error correction (GEC)~\cite{lee2006automatic}, OCR error correction~\cite{tong1996statistical} and spell checking~\cite{fossati2007mixed}. With the recent success of sequence-to-sequence (seq2seq) learning~\cite{sutskever2014sequence} and the transformer model~\cite{vaswani2017attention}, most LST tasks have been tackled by directly training the transformer-based models in a seq2seq fashion. 

While the conventional seq2seq paradigm suits well for general sequence transduction problems such as machine translation, their left-to-right auto-regressive decoding scheme cannot access the future predictions on the right side, which does not fully utilize the characteristic of LST tasks and has been demonstrated to degrade the performance of seq2seq modes~\cite{zhang2018asynchronous,zhang2019synchronous}. 


\begin{figure*}
    \centering
    \includegraphics[width=0.8\textwidth]{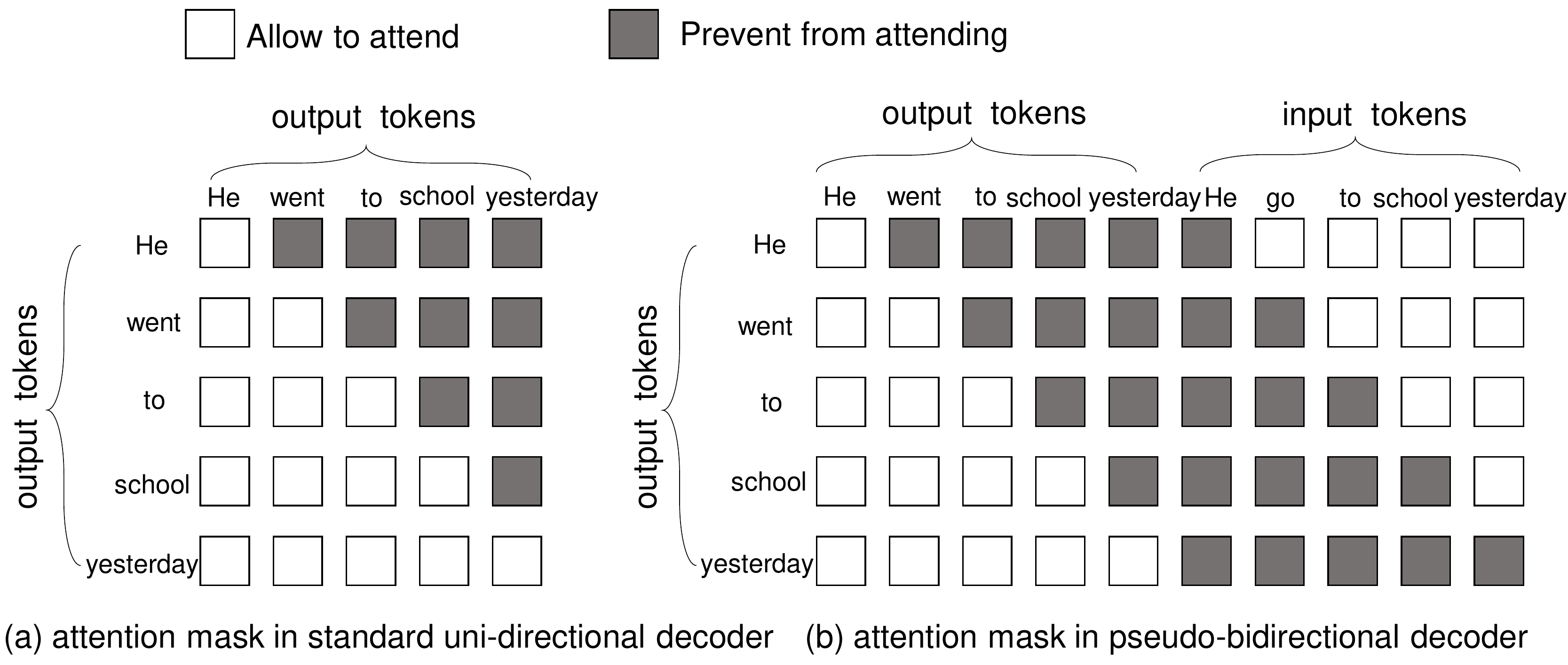}
    \caption{Illustration of the proposed pseudo future modeling approach and the pseudo-bidirectional attention mask used for parallel training. }
    \label{mask}
\end{figure*}

In this work, motivated by the characteristic of LST tasks, we propose a Pseudo Bidirectional Decoding (PBD) approach to tackle LST tasks. Our approach copies the input tokens on the right side of the current decoding position as a proxy for the future tokens. In this way, we augment the decoder of the conventional transformer by allowing it to attend to the representation of ``pseudo'' future tokens in the decoder, making the decoding self-attention module bidirectional without introducing any computational overhead during inference. To retain the parallelizability of the training transformer models, we propose a novel masking strategy that enables the decoder to attend to copied future token representations during training in a parallelizable fashion.  Also, we incorporate a segment embedding mechanism to make the decoder aware of whether a token is directly copied from the encoder and represent them differently from the generated tokens.

With the proposed approach, the encoder and the decoder in transformer models for LST tasks receive similar input sequences and both attend to their bidirectional context information, which motivates us to share all their parameters (except encoder-decoder attention). The parameter sharing mechanism allows us to roughly reduce the total number of parameters of the model by half, which is beneficial for real-world applications and makes the training more efficient. It also explicitly models the characteristic of LST tasks and leads to good regularization effects, enhancing the performance of transformer models on LST tasks and allowing us to train deeper models for further improvements. 

We conduct extensive experiments on three LST tasks including grammatical error correction, spell correction, and OCR correction. Experimental results demonstrate that the proposed PBD approach is able to substantially and consistently improve over competitive transformer baselines across all three LST tasks and yield state-of-the-art results on both spell and OCR correction tasks.

\section{Pseudo-Bidirectional Decoding}


\subsection{Pseudo Future Modeling}

In contrast to recent works on bidirectional decoding~\cite{zhang2018asynchronous,zhang2019synchronous} that employs a right-to-left model at the same time and combines their predictions in a post-hoc fashion with sophisticated algorithms, our approach enables the decoder of the seq2seq model to exploit the future context of the generated sequence without having to predict them in the first place. 

Concretely, our method copies the representation of tokens from the $N+1$ th position to the end of the input sequence in the encoder layer to the corresponding decoder layer as pseudo future information when predicting the $N$ th output token. 
For instance, for grammatical error correction, given an input text ``\textit{He go to school yesterday.}'', a conventional left-to-right decoder would probably generate ``\textit{goes}'' at the second decoding step, as the decoder state is ``\textit{He \textunderscore} '', which is likely to be continued with the third person singular form of the verb ``go''. In contrast, with the proposed pseudo-bidirectional decoding scheme, the decoder state becomes ``\textit{He \textunderscore \ to school yesterday.}'', which facilitates the decoder to correctly generate ``went''. 

While ideally the encoder-decoder attention may capture this information, our method makes the decoder self-attention more effective by allowing it to directly attend to future information, which may be complementary to the information captured by the encoder-decoder attention module, leading to better empirical performance.

\paragraph{Pseudo-bidirectional Attention Mask}

A naive implementation of the PBD approach requires us to change the decoder input for each decoding step instead of feeding the entire output sequence into the decoder and use a causal attention mask to ensure the causality of the decoder. This would hinder the transformer model from being trained in parallel, thus makes the training much less efficient.

To address this problem, we propose a novel masking strategy. As illustrated in Figure \ref{mask}, we concatenate the representation of the input sequences to that of the output sequences to form the key and the value in the decoder self-attention module. The pseudo-bidirectional attention mask makes the decoder self-attention bidirectional by allowing the query tokens to attend to pseudo future tokens copied from the encoder, retaining the causality of the decoder and enabling parallel training.

\paragraph{Segment Embedding}

While the characteristic of LST tasks ensures the copied pseudo future tokens to be similar to the expected output tokens, the simple position-wise alignment method may make the pseudo future information contain some noise. Therefore, we propose a simple segment embedding method that enables the decoder to distinguish the copied tokens from the tokens generated by the decoder and represent them differently. 

Similar to BERT~\cite{devlin2018bert}, we add a learned embedding, which indicates whether the token is generated or copied, to each token representation in each decoder layer. Hopefully, this would make the decoder able to distinguish the copied tokens from the tokens generated by the decoder and represent them differently, thus improve its robustness to the noise in the copied future token representations.

\subsection{Parameter sharing}

The encoder and the decoder in conventional transformer models are independently parameterized for two main reasons. First, the inputs for the encoder and the decoder are usually different for general sequence transduction tasks such as machine translation and text summarization. Second, the encoder self-attention module is bidirectional whereas the decoder self-attention is causal (i.e. uni-directional).

The characteristic of LST tasks ensures the inputs for the encoder and the decoder to be roughly the same, and the proposed pseudo-bidirectional decoding method makes both the encoder and the decoder self-attention module to be bidirectional. This motivates us to share all parameters, except that in the encoder-decoder attention module, between the encoder and the decoder. This roughly reduces the number of parameters by half and also provide some regularization effects.

\section{Experiments}

In this section, we conduct experiments on LST benchmarks to validate the effectiveness of our approach. We mainly focus on the grammatical error correction task and also report results on two other LST tasks including spell and OCR corrections. 

\subsection{Grammatical Error Correction}

\paragraph{Datasets}

Following the recent work~\cite{grundkiewicz2019neural,kiyono2019empirical,zhou2019improving} in GEC, the GEC training data we use is the public Lang-8~\cite{mizumoto2011mining}, NUCLE~\cite{dahlmeier2013building}, FCE~\cite{yannakoudakis2011new} and W\&I+LOCNESS datasets~\cite{bryant2019bea,granger1998computer}. To investigate whether our approaches can yield consistent improvement in this setting, we pretrain our models with 30M sentence pairs obtained by the corruption-based approach and 30M pairs by the fluency boost back-translation approach \cite{ge2018fluency} for GEC pre-training.


\paragraph{Models}

We use the ``transformer-big'' architecture as our baseline model, denoted by \textbf{Transformer}. For throughout comparison, we train two model variants with our approach. The first model (\textbf{Ours}) consists of the same number (i.e. 6) of transformer blocks with the baseline model, thus has the same inference latency while containing only half the number of parameters. The second model is denoted by \textbf{Ours-12 layers}, which consists of 12 transformer blocks, thus has approximately the same number of parameters but the inference latency is $1.7\times$ longer. For comparison, we also train a variant of the ``transformer-big'' architecture with 12 blocks, which is denoted by \textbf{Transformer-12 layers}. For reference, we also compare with a recent model specifically designed for local sequence transduction tasks, denoted by \textbf{PIE}.


We use synthetic data for pre-training and then use the GEC training data to fine-tune the pre-trained models.  The details of model training are provided in the Appendix due to space constraints.


\begin{table}[!t]
	\begin{center}
	\resizebox{1.\linewidth}{!}{
			\begin{tabular}{lccc}
				\hline\hline
				\textbf{Method} & \textbf{\# Parameters}  & \textbf{BEA-19} & \textbf{CoNLL-14} \\ \hline
				\bf PIE (with pretraining) & 345M & - & 59.7  \\ \hline
				\multicolumn{4}{c}{\textbf{w/o Pretraining}} \\ \hline
				\bf Transformer & 210M & 57.1 & 51.5  \\
				\bf Transformer-12 layers & 383M & 56.3 & 51.3  \\
				\bf Ours & 132M & 58.6 &  53.7  \\
				 \multicolumn{2}{l}{~-w/o future modeling} & 57.6 & 51.8 \\
				\multicolumn{2}{l}{~-w/o parameter sharing} & 58.2 & 52.9 \\ 
				\bf Ours-12 layers & 232M & \bf 59.5 & \bf 54.4 \\ 
				\multicolumn{2}{l}{~-w/o future modeling} & 58.6 & 52.1 \\
				\multicolumn{2}{l}{~-w/o parameter sharing} & 58.8 & 53.8 \\ \hline
\multicolumn{4}{c}{\textbf{w/ pretraining}} \\ \hline
				\bf Transformer & 210M & 61.2 & 57.1  \\
				\bf Transformer-12 layers & 383M  & 61.9 & 57.5  \\
				\bf Ours  & 132M & 63.2 & 58.9 \\
				 \multicolumn{2}{l}{~-w/o future modeling} & 61.5 & 57.4 \\
				\multicolumn{2}{l}{~-w/o parameter sharing} & 61.7 & 57.9 \\ 
				\bf Ours-12 layers & 232M & \bf 63.9 & \bf 60.1 \\
				 \multicolumn{2}{l}{~-w/o future modeling} & 61.8 & 57.7 \\
				\multicolumn{2}{l}{~-w/o parameter sharing} & 62.1 & 58.2 \\ 
				\hline\hline
		\end{tabular}}
	\end{center}
	\caption{\label{gecresult} The performance of different compared models on two test sets of GEC task.}
\end{table}

\paragraph{Evaluation}

We evaluate the performance of GEC models on the BEA-19 and the CoNLL-14 benchmark datasets. 
Following the latest work in GEC~\cite{grundkiewicz2019neural,kiyono2019empirical,zhou2019improving, zhang2019sequence}, we evaluate the performance of trained GEC models using $F_{0.5}$ on test sets using official scripts\footnote{M2scorer for CoNLL-14; Errant for BEA-19.} in both datasets. 

\paragraph{Results}

The performance of different compared models on the GEC task is shown in Table \ref{gecresult}.  Note that we only compare against transformer models with the same pretraining/fine-tuning data in our setting for fair comparison as our contribution is orthogonal to better data synthesis method.

We can see that for the same configuration of the ``transformer-big'' baseline, our approach outperforms the baseline by a large margin in both settings with and without pretraining with synthetic data. This suggests that our approach is able to improve the performance of transformer-based LST models while reducing the number of parameters by half. In addition, we can see that by doubling the number of transformer blocks, our model is able to yield substantial further improvement. In contrast, we can see that simply increasing the number of transformer blocks (i.e. \textbf{Transformers-12 layers}) fails to improve the performance. This implies that our approach may be able to facilitate the training of deeper transformer models by providing regularization effects. 


We then conduct an ablation study where either the pseudo future context modeling approach or the parameter sharing mechanism is disabled to better understand their relative importance. The results are shown in Table \ref{gecresult}. We can see that the proposed pseudo future modeling approach method is very important in both the default and the deeper configuration of our transformer-based models, demonstrating its effectiveness on LST tasks. We also find that the parameter sharing mechanism is more effective in deeper models. This suggests that the parameter sharing mechanism may provide strong regularization effects and make it easier to train deeper transformer models.

\subsection{More Sequence Transduction Tasks}

Following previous work~\cite{ribeiro2018local,awasthi2019parallel}, we demonstrate the effectiveness of the proposed approaches on two additional local sequence transduction tasks including spell and OCR correction. We employ a two-layer transformer and a four-layer transformer as the backbone model for comparison and evaluate the compared models on the twitter spell correction datasetand the Finnish OCR dataset described as follows:

\paragraph{Spell correction}

We use the twitter spell correction dataset~\cite{aramaki2010typo} which consists of 39172 pairs of original and corrected words obtained from twitter. We use the same train-dev-valid split as~\cite{ribeiro2018local} (31172/4000/4000). We tokenize on characters and our vocabulary comprises the 26 lower cased letters of English.

\paragraph{OCR correction}

We use the Finnish OCR data set3 by~\cite{silfverberg2016data} comprising words extracted from Early Modern Finnish corpus of OCR processed newspaper text. We use the same train-dev-test splits as provided by~\cite{silfverberg2016data}. We tokenize on characters and our vocabulary comprises all the characters seen in the training data.

\begin{table}[!t]
	\begin{center}
			\begin{tabular}{lcc}
				\hline\hline
				\textbf{Method} & \textbf{Spell} & \textbf{OCR} \\ \hline
				 \bf LSTM-soft &  46.3 &  79.9  \\
				 \bf LSTM-hard &  52.2 &  58.4  \\
				 \bf \citet{ribeiro2018local} &  54.1 &  81.8  \\
				 \bf PIE &  67.0 &  87.6  \\ \hline
				\multicolumn{3}{c}{\textbf{2 Layers}} \\ \hline
				\bf Transformer & 67.6 & 84.5  \\
				\bf Ours & \bf 69.2 & \bf 88.7  \\ \hline
    \multicolumn{3}{c}{\textbf{4 Layers}} \\ \hline
                \bf Transformer-4 layers & 67.1 & 85.4  \\
				\bf Ours-4 layers & \bf 70.4 & \bf 89.6 \\
				\hline\hline
		\end{tabular}
	\end{center}
	\caption{\label{otherlst} The performance (accuracy) of different compared models on the spell and OCR correction tasks.}
\end{table}



\paragraph{Results}

The result is shown in Table \ref{otherlst}. We can see that the proposed method is able to significantly outperform the vanilla transformer-based models, as well as the LSTM and sequence labeling based LST baselines in both settings where either the number of parameters in the model is the same or the inference latency is the same, which is consistent with the result in the GEC task. Our deeper model variant yields the state-of-the-art results in both tasks. This demonstrates the effectiveness of the proposed approach and suggests that our approach is versatile for different LST tasks.

\section{Related work}
\paragraph{Local Sequence Transduction}~\citet{ribeiro2018local} proposed to formulate LST tasks as sequence labeling problems by first predicting insert slots in the input sequences using learned insertion patterns and then using a sequence labeling task to output tokens in the input sequences or a special ``delete'' token. \citet{awasthi2019parallel,malmi-etal-2019-encode} propose to predict output edit operations including word transformations and further improve the performance of sequence labeling based LST models. Their approaches require massive engineering efforts to design an appropriate set of word transformations, which makes it non-trivial to generalize to other LST tasks. Also, the sequence labeling formulation lacks the flexibility of seq2seq models because it can only make local edits, which is demonstrated by their inferior performance.  More recently, \citet{li2020towards} use BERT to perform local sequence transduction with the method proposed by \citet{zhou2019bert}. However, their method mainly suits for the cases where the length of the output sentence is unchanged.


\paragraph{Bidirectional Decoding and Future Modeling} Previous work~\cite{sennrich2016edinburgh,deng2018alibaba} investigate using right-to-left models to re-rank the generated sentences. More recently, \citet{xia2017deliberation} and \citet{zheng2018modeling} propose two-pass decoding to model the right-side information, while~\citet{zhang2018asynchronous,zhang2019synchronous} use bidirectional beam search algorithms to generate the output sequences. These approaches integrate the right side context indirectly and introduce substantial computational overhead during inference, which is undesirable for real-world applications. 


\paragraph{Parameter Sharing in Transformer} Several parameter sharing mechanisms have been explored in transformer-based models. ALBERT~\cite{lan2019albert} shares all encoder layers to reduce the number of parameters in the pretrained language model. \citet{xia2019tied} propose to share the encoder and the decoder in transformer-based machine translation models. The performance gain in their setting is relatively small, which may be due to the discrepancy in the input sequences and the attention direction in the encoder and the decoder. 


\section{Conclusion}

In this paper, motivated by the characteristic of local sequence transduction tasks, we propose pseudo-bidirectional decoding (PBD) to provide approximated future information for transformer-based LST models and share the parameters between the encoder and the decoder of LST models to provide regularization effects while reducing the number of parameters. Our experiments on three LST tasks shows that our approach is able to yield consistent improvements upon strong transformer baselines while significantly reducing the number of parameters in the model. 

\section*{Acknowledgments}
We thank the anonymous reviewers for their valuable comments.


\bibliographystyle{acl_natbib}
\bibliography{emnlp20}

\end{document}